\documentclass[10pt,twocolumn,letterpaper]{article}

\usepackage{wacv}
\usepackage{times}
\usepackage{epsfig}
\usepackage{graphicx}
\usepackage{amsmath}
\usepackage{amssymb}     
\usepackage{adjustbox}
\usepackage{graphics} % for pdf, bitmapped graphics files
\usepackage{epsfig} % for postscript graphics files
\usepackage{mathptmx} % assumes new font selection scheme installed
\usepackage{times} % assumes new font selection scheme installed
\usepackage{amsmath} % assumes amsmath package installed
\usepackage{amssymb}  % assumes amsmath package installed

\wacvfinalcopy % *** Uncomment this line for the final submission

\def\wacvPaperID{****} % *** Enter the wacv Paper ID here

\ifwacvfinal\fi
\begin{document}

%%%%%%%%% TITLE
\title{Interactive Generative Adversarial Networks for Facial Expression Generation in Dyadic Interactions}

% Authors at the same institution
%\author{First Author \hspace{2cm} Second Author \\
% Authors at different institutions
\author{Behnaz Nojavanasghari \\
University of Central Florida\\
{\tt\small behnaz@eecs.ucf.edu}
\and
Yuchi Huang \\
Educational Testing Service\\
{\tt\small yhuang001@ets.org }
\and
Saad Khan \\
Educational Testing Service\\
{\tt\small skhan002@ets.org  }
}

\maketitle
\ifwacvfinal\thispagestyle{empty}\fi

%%%%%%%%% ABSTRACT
\begin{abstract}
A social interaction is a social exchange between two or more individuals, where individuals modify and adjust their behaviors in response to their interaction partners. Our social interactions are one of most fundamental aspects of our lives and can profoundly affect our mood, both positively and negatively. With growing interest in virtual reality and avatar-mediated interactions, it is desirable to make these interactions natural and human like to promote positive effect in the interactions and applications such as intelligent tutoring systems, automated interview systems and e-learning. In this paper, we propose a method to generate facial behaviors for an agent. These behaviors include facial expressions and head pose and they are generated considering the users affective state. Our models learn semantically meaningful representations of the face and generate appropriate and temporally smooth facial behaviors in dyadic interactions.  
\end{abstract}
%%%%%%%%%%%%%%%%%%%%%%%%%%%%%%%%%%%%%%%%%%%%%%%%%%%%%%%%%%%%%%%%%%%%%%%%%%%%%%%%

\section{Introduction}
Designing interactive virtual agents and robots have gained a lot of attention in recent years \cite{fong2003survey, kanda2004interactive, ulicny2001crowd}. The popularity and growing interest in these agents is partially because of their wide applications in real world scenarios. They can be used for variety of applications from education \cite{nojavanasghari2016emoreact,nojavanasghari2016future}, training \cite{rickel2001intelligent} and therapy \cite{dautenhahn2004towards,nojavanasghari2017exceptionally} to elderly care \cite{broekens2009assistive} and companionship \cite{{dautenhahn2006may}}. One of the most important aspects in human communication is social intelligence \cite{bar2004emotional}. A part of social intelligence depends on understanding other people's affective states and being able to respond appropriately. To have a natural  human- machine interaction, it is critical to enable machines with social intelligence as well. The first step towards this capability is 
%%%%%%%%%%%%%%%%%%%%%%%%%%%%%%%%%%%%%%%%
   \begin{figure}[ht!]
      \centering
      \includegraphics[scale=0.7]{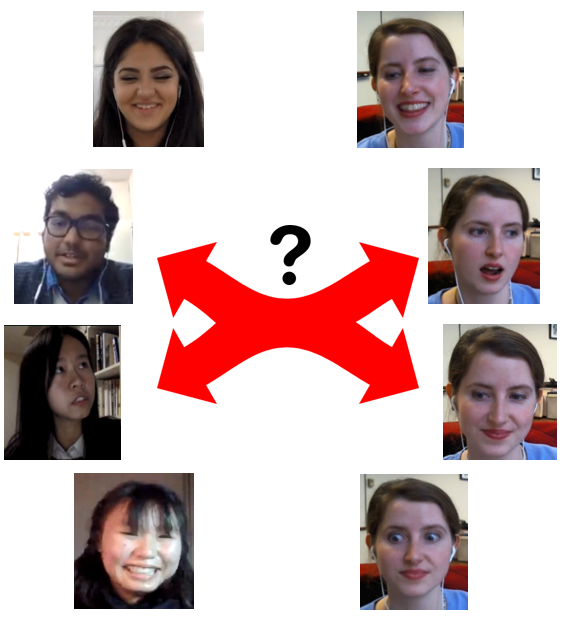}
      \caption{The problem we are solving in this paper is, given affective state of one person in a dyadic interaction, we generate the  facial behaviors of the other person. By facial behaviors, we refer to facial expressions, movements of facial landmarks and head pose.}
      \label{figurelabel}
   \end{figure}
%%%%%%%%%%%%%%%%%%%%%%%%%%%%%%%%%%%%%%%%
observing human communication dynamics and learning from the occurring behavioral patterns \cite{nojavanasghari2017hand2face,nojavanasghari2016deep}. In this paper, we focus on modeling dyadic interactions between two partners. The problem we are solving is given the affective state of one partner, we are interested in generating non-verbal facial behaviors of the other partner. Figure 1 shows the overview of our problem. The affective states that we consider in generating facial expressions include joy, anger, surprise, fear, contempt, disgust, sadness and neutral. The non-verbal cues that we are generating in this work, are  facial expressions and head pose. These cues help us generate head nods, head shakes, tilted head and various facial expressions such as smile which are behaviors that happen in our daily face to face communications. A successful model should learn meaningful factors of the face for generating of most appropriate responses for the agent. We have designed our methodology based on the following factors: 1) humans react to each others' affective states from an early age and adjust their behavior accordingly \cite{tronick2009still}. 2) Each affective state is characterized by particular facial expressions and head pose\cite{ekman1992argument}. 3) Each affective state can have different intensities, such as happy vs very happy. These factors (e.g. affective state and their intensities) effect how fast or slow or facial expressions and head pose changes over time\cite{ekman1992argument}. 

There has been great deal of research on facial expression analysis where the effort is towards describing the facial expressions \cite{tian2011facial}. Facial expressions are a result of movements of facial landmarks such as raising eyebrows, smiling and wrinkling nose. A successful model for our problem, should learn the movements of these points in response to different affective states.  

We are using FACET SDK for extracting 8-dimensional affective state of one partner in the interaction. We then use it as a condition in generation of non-verbal facial behaviors of the other person. Our generation process relies on a conditional generative adversarial networks \cite{mirza2014conditional}, where the conditioning vector is affective state of one party in the interaction. Our proposed methodology is able to extract meaningful information about the face and head pose and integrate the encoded information with our network to generate expressive and temporally smooth face images for the agent.

\section{Related work}

Humans communicate using not only words but also non-verbal behaviors such as facial expressions, body posture and head gestures. 
In our daily interactions, we are constantly adjusting our non-verbal behavior based on other peoples behavior. In collaborative activities such as interviews and negotiations, people tend to unconsciously mimic other people's behavior \cite{bilakhia2013audiovisual, lakin2003chameleon}. 
%%%%%%%%%%%%%%%%%%%%%%%%%%%%%%%%%%%%%%%%%%%%%%%%%%%
\begin{figure*}[ht!]
      \centering
      \includegraphics[width=\linewidth]{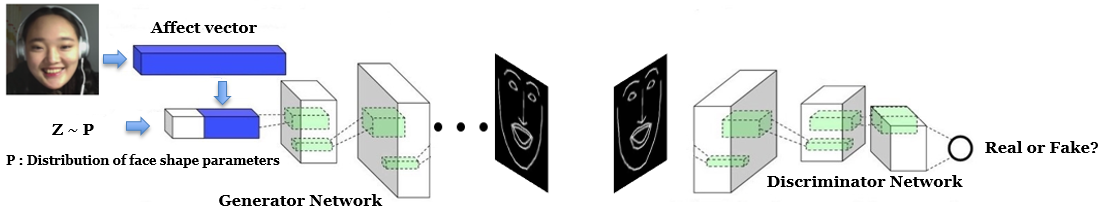}
      \caption{Overview of our affect-sketch network. In first stage of our two-stage facial expression generation, we generate face sketches conditioned on the affective state of the interviewee and using a z vector that carries contains semantically meaningful information about facial behaviors. This vector is sampled from generated distributions of rigid and non-rigid face shape parameters using our proposed methods.}
      \label{figurelabel}
   \end{figure*}

   \begin{figure}[ht!]
      \centering
      \includegraphics[width=\linewidth]{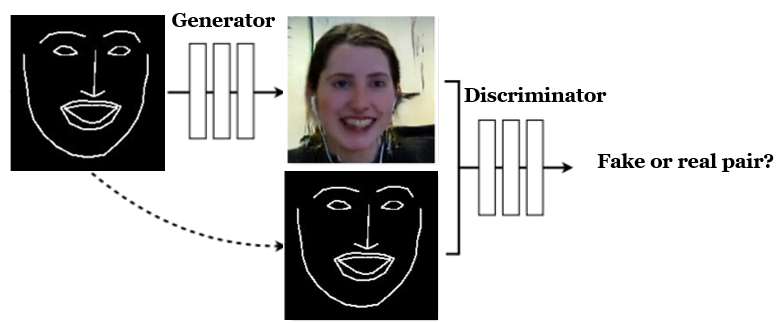}
      \caption{Overview of our sketch-image network. In the second stage of our two-stage facial expression generation, we generate face images conditioned on the generator sketch of the interviewer which is generated in the first stage of our network. }
      \label{figurelabel}
   \end{figure}
   
   %%%%%%%%%%%%%%%%%%%%%%%%%%%%%%%%%%%%%
With increasing interest in social robotics and virtual reality and their wide applications in social situations such as automated interview systems \cite{hoque2013mach}, companionship for elder care \cite{broekens2009assistive} and therapy for autism \cite{dautenhahn2004towards}, there is a growing need to enable computer systems to understand, interpret and respond to people's affective states appropriately. 

As a person's face discloses important information about their affective state, it is very important to generate appropriate facial expressions in interactive systems as well. 

There has been a number of studies focusing on automatic generation of facial expressions. These efforts can be categorized into three main groups: 1) Rule based approaches where affective states are mapped into a pre-defined 2D or 3D face model \cite{heylen2001generation}. 2) Statistical approaches where face shape is modeled as linear combination of prototypical expression basis \cite{lisetti2000automatic}. 3) Deep belief networks where the models learn the variation of facial expressions in presence of various affective states  and produce convincing samples \cite{susskind2008generating}.

While generating facial expressions using previous approaches can result in promising  results, it is important to consider other people's affective states while generating facial expressions in dyadic interactions. There has been previous work in this area, \cite{huang2017dyadgan} where researchers use conditional generative adversarial networks for generating facial expressions in dyadic interactions. While the proposed approach results in appropriate facial expression for one frame, it does not consider the temporal consistency in a sequence, which results in generating non-smooth facial expressions over time. Also, some of the attributes such as head position and orientation that are important cues for recognizing tilted head, head nods, head shakes are not captured by the model. 

In this paper, we address the problem of generating facial expressions in dyadic interactions while considering temporal constraints and additional facial behaviors such as head pose. Our proposed model extracts semantically meaningful encodings of the face regarding movements of facial landmarks and head pose and integrate it with the network which results in generation of appropriate and smooth facial expressions.
%%%%%%%%%%%%%%%%%%%%%%%%%%%%%%%%%%%%%%%%%%%%%%%%%%%%%%%%%%%%%%%%%%%%%%%%%%%%%%%%

\section{Methodology}

Generative adversarial networks (GANs) \cite{goodfellow2014generative} have been widely used in the field of computer vision and machine learning for various applications such as image to image translation \cite{isola2016image}, face generation \cite{gauthier2014conditional}, semantic segmentation \cite{luc2016semantic} and etc. GANs are a type of generative model which learn to generate based on generative and discriminative networks that are trained and updated at the same time. The generative network tries to find the true distribution of the data and generate realistic samples and the discriminator network tries to discriminate between the samples that are generated by generated and samples from real data. The general formulation of conditional GANs are as follows:
\begin{multline} 
\min_{G} \max_{D}  \mathcal{L}(D, G)= \mathbb{E}_{x,y \sim p_{data}(x,y)} [\log D(x,y)] + \\
\mathbb{E}_{x \sim p_{data}(x), z \sim p_{z} (z)} [\log (1 - D(x,G(x, z)))]. \label{GANs}
\end{multline}
Conditional GANs are generative models that learn a mapping from random noise vector z to output image y conditioned on auxiliary information $x: G : \{x, z\} \Rightarrow y$. A conditional GAN consists of a generator G(x, z) and a discriminator D(x, y) that compete in a two-player minimax game: the discriminator tries to distinguish real training data from generated images, and the generator tries to fail the discriminator. That is, D and G play the following game on V (D, G). As you can see the goal of these networks is to increase the probability of samples generated by generative networks to resemble the real data so that the discriminator network fail to differentiate between the generated samples and the real data.  

Conditional generative adversarial networks (CGANs) are a type of GAN that generate samples considering a conditioning factor. As our problem is generation of facial expressions conditioned on affective states, we found CGANs as perfect candidate for addressing our problem. We chose to use CGANs to generate facial expressions of one party, conditioned on the affective state of the other partner in a dyadic interaction.The GAN formulation uses a continuous input noise vector z, that is an input to the generator. However, this vector is a source of randomness to the model and does not capture semantically meaningful information. However, using a meaningful distribution which corresponds to our objective and desired facial behaviors, can help the model integrate meaningful information with the network and improve the results of generation. For example, in case of generating facial behaviors, the important factors are the variation in locations of facial landmarks such as smiling and head pose such as tilted head.  

  \begin{figure}[ht!]
      \centering
      \includegraphics[scale = 0.8]{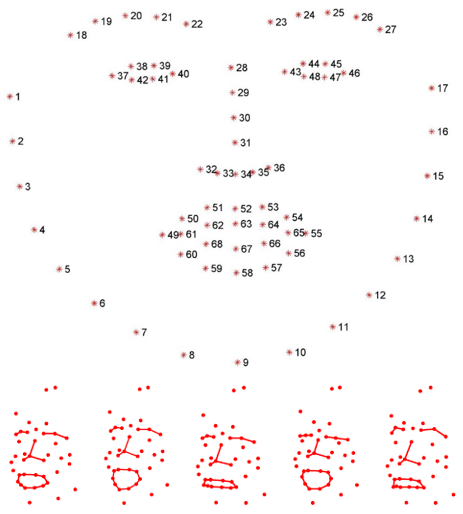}
      \caption{This figure shows 68 landmarks of the face that we intend to generate using our model. In the bottom row you see a visualization of non-rigid shape parameters which correspond to movements of facial landmarks such as opening the mouth. Our model aims to learn the distribution of these parameters and use them in generating temporally smooth sequences. }
      \label{figurelabel}
   \end{figure}

In this paper, we present a modification to the sampling process of the latent vector z which is an input to the CGAN network. We sample z  from the meaningful distributions of facial behaviors that are generated by two different approaches 1) Affect-shape dictionary and 2) Conditional LSTM. This step helps the network to learn interpretable and meaningful representation of facial behaviors. In the next sections we will first explain the two stage architecture of our network. Then we will explain preprocessing of the frames by going over the intuition behind our choice of face representation. Finally, we provide detailed description of our methods for generating distributions for vector z which is the input to our CGAN networks.  
\subsection{Two-Stage Conditional Generative Adversarial Network}
To generate facial behaviors for agent, we use a two stage architecture, both of which are based on conditional generative adversarial networks. Figure 2 shows an overview of our first network. In the first stage we use a conditional generative adversarial network which takes face shape parameter as input  for vector z as well as the conditioning vector which is the 8 affective states of one partner in the interaction, and it generates sketches of faces for the other partner in the interaction. Note that the z vector will be sampled using one of our proposed strategies, which are explained in the following sections. In the second stage, we input the generated sketches in the first stage as input to GAN and generate the final facial expressions. Figure 3 shows an overview of our second network.
   \begin{figure*}
      \centering
      \includegraphics[scale = 0.6]{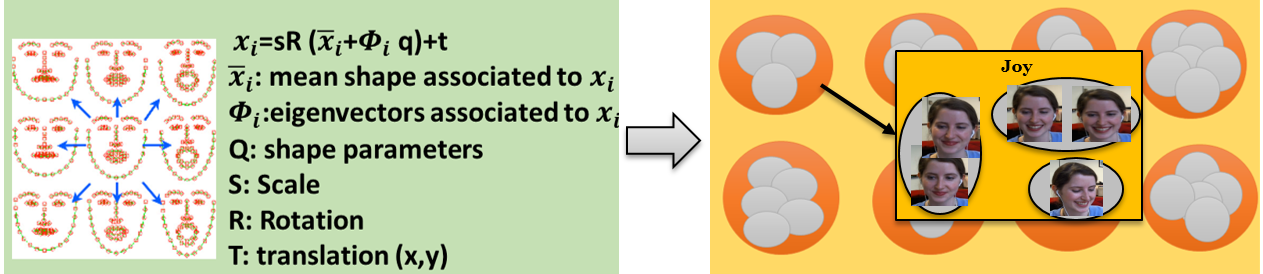}
      \caption{Overview of our affect-shape approach for generating clusters of face shape parameters. We first extract shape descriptors for all faces in the dataset. Then we cluster them into 8 affect classes corresponding to Joy, Anger, Surprise,
Fear, Contempt, Disgust, Sadness and Neutral. Then, we do a further inter class clustering for each affect class. The purpose of the second clustering is to find the inter affect clusters, such as very joyful, joyful or little joyful.These clusters form the distributions from which the z vector is sampled in our generative models. }
      \label{figurelabel}
   \end{figure*}
   \begin{figure*}
      \centering
      \includegraphics[width=\linewidth]{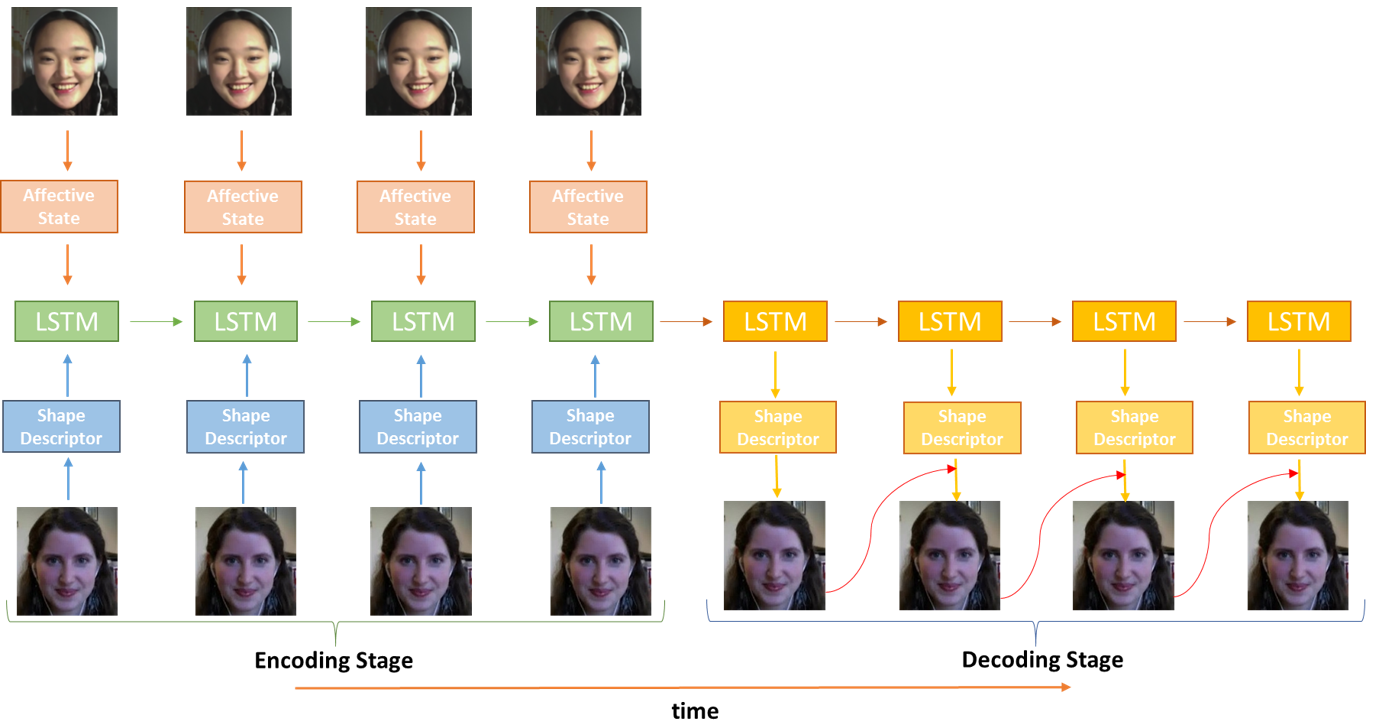}
      \caption{Overview of our conditional LSTM approach for generating facial shape parameters. In this approach we consider the affective state of the interviewee and past frames of the interviewer's facial expressions as conditioning vector and previous history. We then learn to generate the future frames of interviewer's facial expressions using a conditional LSTM. The output result of this generation is the input for z vector in our generative models. }
      \label{figurelabel}
   \end{figure*}

\subsection{Data Preprocessing and Face Representation}

Facial landmarks are the salient points on face located at the corners, tips or midpoints of facial components such as eyes, nose, mouth, etc \cite{tie2013automatic}. Movements of these points form various facial expressions that can convey different affective states. For example, widening the eyes can communicate surprise and a smile can be an indication of happiness. Since the locations of facial landmarks alone are not particularly meaningful in communicating affect but rather their movements and change over time which plays a rule in affect recognition \cite{senechal2011facial}, we decided to measure their movements and change over time. To do so, we are projecting a 3D point distribution model on each image. This model has been trained on in the wild data \cite{koestinger2011annotated}. The following equation is used to place a single feature point of the 3D PDM in a given input image:
\newline
\centerline{
$\bar{x}_{i}=s.R_{2D}.(\bar{X}_{i}+\phi_{i}q)+t$}

In this equation, s shows scale, R shows the head rotation and t is the head translation which are known as rigid- shape parameters. $\bar {X}_{i} = [\bar{x}_{i},\bar{y}_{i},\bar{z}_{i}]$ is the mean value of $i$ $th$ feature , $\phi _{i}$ is principle component matrix and q is a matrix controlling the non-rigid shape parameters, which correspond to deviations of landmarks locations from an average (neutral) face. 
\newline
As described, shape parameters of the face carry rich information about head pose and facial expressions. Hence, we use them as our face representations.  Figure 4 shows 68 facial landmarks that we have considered in this work. We have also included a visualization of non-rigid shape parameters at the bottom row for better understanding of these parameters.

\subsection{Affect-Shape Dictionary}

Different facial expressions are indicative of different affective states \cite{keltner2003facial}. Depending on the affective state and its intensity, the temporal dynamics and changes in landmarks can happen in various facial expressions such as a smile vs a frown and they can happen slow or fast depending on how intense is the affect \cite{decety2014neural}.  Based on this knowledge, we propose the following strategies to learn meaningful distributions of face shape parameters. Figure 5 shows an overview of our affect-shape dictionary generation. 
\subsubsection{Affect Clustering}
While in state of fear landmarks can change very sudden, in state of sadness or neutral the changes in locations of landmarks will be smaller and slower. In order to account for this point, we first cluster the calculated shape parameters for all input frames into 8 clusters, using k-means clustering \cite{hartigan1979algorithm} where clusters denote the following affective states: Joy, Anger, Surprise,
Fear, Contempt, Disgust, Sadness and Neutral. This stage gives us distributions of possible facial expressions for each affective state. 
\subsubsection{Inter Affect Clustering}
The output of the first clustering step will give us  groupings of shape parameters in 8 clusters corresponding to each category of affective states. Each affective state can have various arousal levels (a.k.a intensity)\cite{kensinger2004two}. We did a second level of clustering of shape parameters inside each affect cluster. To do so, we used a hierarchical agglomerative clustering approach \cite{franti2006fast}, where we put the constraint that each cluster should contain at least 100 videos. The result of this clustering gives us 3 to 9 sub clusters corresponding to different arousal levels of each affective state. Having information about the possible facial expressions that can happen in a particular affect category and what has been expressed in previous frames by the agent, we can sample z from an appropriate distribution which results in an expressive and temporally smooth sequence of facial expressions. To account for the temporal constraints, for generating $n th $ frame, we pick z  from the appropriate distribution based on the affective state which is the closest to the previously chosen z in terms of euclidean distance. 

\subsection{Conditional LSTM}
Long Short Term Memory networks (LSTM) have been known for their ability in learning long-term dependencies \cite{hochreiter1997lstm}. We are interested in learning the dynamics of facial expression, hence LSTMs are an appropriate candidate for our problem. As the problem we are solving is a conditional problem, we will use a conditional LSTM (C-LSTM) \cite{augenstein2016stance}.  The input to CLSTM will be the concatenated vector of affective state of the interviewer and the shape parameters of the interviewee for $n$ previous frames and the generated output will be the future shape parameters of the interviewee. We fix $n$ to be 100 frames in our experiments. We use conditional LSTMS in two different settings which are described in the following subsections. Figure 6 shows an overview of our C-LSTM approach.

\textbf{Overlapping C-LSTM:}
In overlapping conditional LSTMS, we consider information across 100 frames as input of the model and the output of the model is generation of shape parameters for one future frame. In each step, the newly generated frame is added to the history and the information from the first input frame is removed instead ( in first in, first out manner) , to predict futures frames. 

\textbf{Non-overlapping C-LSTM:}
In non-overlapping C-LSTMs the input to the models is information across 100 frames and the output is generation of shape parameters for future 100 frames, with no overlap between the frames.

\begin{figure*}[ht!]
\centering \includegraphics[width=\linewidth]{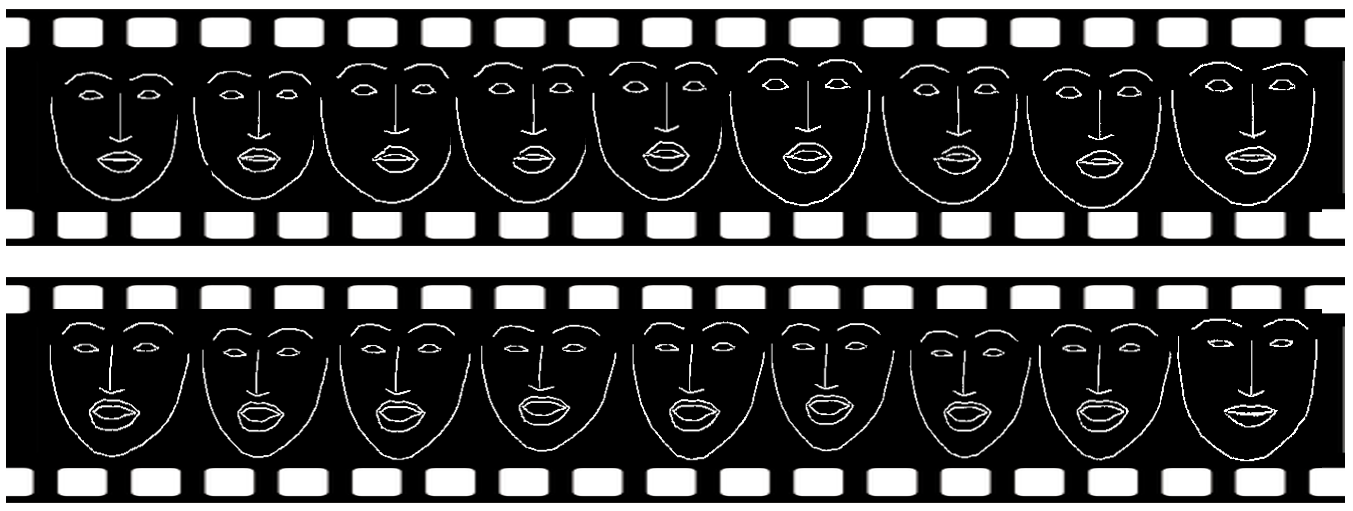}
      \caption{Examples of generated sketches using both of our approaches. The first sequence corresponds to the result of affect-shape dictionary approach and the second row is generated using the conditional LSTM approach. Overall, our first approach generates smoother sequences. }
      \label{figurelabel}
   \end{figure*}

\section{Experiments}
To assess the effectiveness of our proposed approaches in generating appropriate and temporally smooth facial expressions, we performed a number of experiments using the proposed frameworks approaches. Our experiments show the
comparison between our models in terms of result and their pros and cons. 
\subsection{Dataset}
The dataset used in this paper is 31 pair of dyadic interactions (videos) which are interviews for undergraduate admissions process. The purpose of interviews is to assess English speaking ability of the prospective college. There are 16 male and 15 female candidates and each candidate is interviewed by the same interviewer (Caucasian female) who followed a predetermined set of academic
and nonacademic questions designed to encourage open conversation. The interviews were conducted using Skype video conferencing so the participants could see and hear each other and the video data from each dyadic interaction was captured. The duration of interviews varies from 8 to 37 minutes and a total of 24 hours of video data. We have used 70,000 short video clips for training data and 7000 videos for evaluation.
\subsection{Affect Recognition Framework}
For estimating the affective state of the interviewee, we have used Emotient’s Facet SDK \cite{facet} to process the frames and estimate 8- dimensional affect descriptor vectors, representing the likelihood of the following classes: joy, anger, surprise, fear, contempt, disgust, sadness and neutral. 
\subsection{Implementation Details}
To create our training set we randomly sampled 70,000 video clips of 100 frames each (∼3.3 seconds) from the interviewee videos and extracted their affect descriptor using Facet SDK. These affect descriptors are then combined into a single vector using the approach proposed by Huang et al. \cite{huang2017dyadgan}. For each interviewee video clip a single frame from the corresponding interviewer video is also sampled for training. Our approach is to train the model to generate a single frame of the interviewer conditioned on facial expression descriptors from the preceding 100 frames of the interviewee. All face images are aligned and landmarks are estimated by the method proposed by kazem et al. \cite{kazemi2014one} to generate ground truth face sketches. 
For creating our test data, we randomly sampled 7000 interviewee video clips of 400 frames each (no overlap with the training set) and used these as input to our two model to generate 7000 facial expression video clips. 
  
For training the first stage of our network (affect to sketch), we first estimate the landmarks of the interviewer using the proposed method by Kazemi et al. \cite{kazemi2014one} which gives us 68 facial landmarks (See Figure 4). We then connect these points by linear lines of one pixel width to generate the sketch images. The generated sketches are used in a pair with the corresponding image of the interviewer for training the second network. We masked all training image pairs by an oval mask after roughly aligned all faces, to reduce the effect of appearance and lightening variation of the the face in generating our videos. In the generator, a sketch image is passed through an encoder-decoder network \cite{isola2016image} each containing 8 layers of down sampling and up sampling, in order to generate the final image. We have adopted the idea proposed by Ronneberger et al. \cite{ronneberger2015u} and have connected feature maps at layer i and layer n-i where n is total number of layers in the network. In this network receptive fields after convolution are concatenated with the receptive fields in up-convolving process, which allows network to use original features in addition to features after up-convolution and results in overall better performance than a network that has access to only features after up-sampling.
%%%%%%%%%%%%%%%%%%%%%%%%%%%%%%%%%

   \begin{figure*}[ht]
\centering \includegraphics[width=\linewidth]{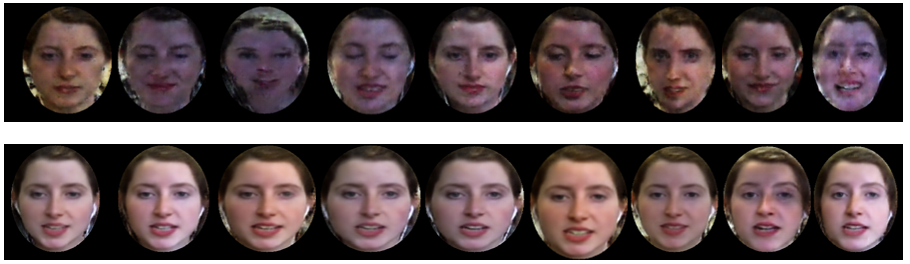}
      \caption{Examples of generated face images using the second stage of our architecture. The first sequence corresponds to a single step generation without considering the sketches and generating face images directly from the affect vector and noise. The second row shows the results related to generation of faces using our proposed two stage architecture that we have used in our experiments.  Note that the quality of images in the second row is much better than the first row. }
      \label{figurelabel}
   \end{figure*}

%%%%%%%%%%%%%%%%%%%%%%%%%%%%%%%%%%%%%%%%%%%%%%%%%%%%%%%%%%%%%%%%%%%%%%%%%%%%%%%%
We adopted the architecture proposed by Radford et. al \cite{radford2015unsupervised} for our generation framework and deep convolutional structure for generator and discriminator. We used modules of the form convolution-BatchNorm-ReLu \cite{ioffe2015batch} to stabilize optimization. In the training phase, we used mini-batch SGD and applied the Adam solver. To avoid the fast convergence of discriminators, generators were updated twice for each discriminator update, which differs from original setting \cite{radford2015unsupervised} in that the discriminator and generator update alternately.

In order to generate temporally smooth sequences, for each new frame, z is sampled from a meaningful distribution and based on the previous frames.  Note that the appropriateness of z for a frame is decided based on the possible variation between two adjacent frames in distribution of face shape parameters training data which correspond to the movements of facial landmarks and head pose.

\subsection{Results and Discussions}
We generated sequences of non-verbal facial behaviors for an agent, based on affect-shape strategy and conditional LSTM. Figure 7 shows the output sketches generated by these two models. The first row corresponds to the affect-shape dictionary approach and the second row shows the results based on Conditional LSTM. We observed that the results from dictionary based is more smooth. In results of conditional LSTM there seems to be grouping of smooth sequences followed by sudden jumps between the sequences (In Figure 7 see the transition between last and one before last frame in the second row). We think that this can happen because of number of frames we have considered as history, as in LSTM frameworks all of the history gets summarized and in unrolling step the time steps are not captured accurately and the generation is more based on an average of all frames.
We  compare the two proposed strategies in Table 1. The advantage of the dictionary based approach is that it does not require previous history of the face and it can run in real time. However, since z is randomly sampled from the possible distribution space, it generates a different sequence each time and it is hard to evaluate this approach.

In table 2 we show an evaluation of conditional LSTM approach for generating the facial expressions. We have generated the facial expressions using C-LSTM by considering 100 frames as history and  1) generating the 101 st frame (complete overlap). 2) generating the next 100 frames (No overlap). As this method, tries to generate the closest sequence to ground truth, we have calculated mean squared error as an evaluation metric. 

As you can see, C-LSTM with 100 frame overlap has a smaller error value compared to the C-LSTM with no overlap. Intuitively, it is harder to predict future with no overlap than having overlap. 
\begin{table}[ht!]
\centering
\begin{center}
\begin{adjustbox}{width=0.47\textwidth}
\begin{tabular}{|l||c|c|c|l| }
 \hline
\textbf{Properties}&\textbf{ Affect-Shape Dictionary}& \textbf{Conditional LSTM}\\
 \hline
\textbf{History}   & No need for history. & Needs history.\\
\textbf{Diversity}&   Generates diverse sequence.& Generates one particular sequence. \\
\textbf{Evaluation}&  Hard to evaluate. & Easier to evaluate.\\
\textbf{Speed}&Real time& Not real time \\
 \hline\end{tabular}
\end{adjustbox}
\end{center}
\caption{Comparison between Affect-shape dictionary and conditional LSTM approaches in generating face shape parameters. }
\end{table}

\begin{table}[ht!]
\begin{center}
\begin{adjustbox}{width=0.45\textwidth}
 \begin{tabular}{||c c c||} 
 \hline
 Method & C-LSTM w overlap &  C- LSTM w/o overlap \\ [0.5ex] 
 \hline\hline
 MSE & 0.101  & 0.183\\ 
[1ex] 
 \hline
\end{tabular}
\end{adjustbox}
\end{center}
\caption{Comparison between performance of conditional LSTMs with and without overlap in history.}
\end {table}

We have also generated the final face images using our CGAN network. Figure 8 shows the generated images of the interviewer which is the final result of the our second network, sketch-image GAN. The top row shows the results of such images if we only had one network and directly went from affect vector and z into face images. The second row shows the results of a two step network where we first generate face sketches and then face images. Note that quality of the images in second row is significantly better than the first row. Also, as our final goal is to transfer these expressions to an avatar's face, our mid-level results can be used to transfer the expressions to various virtual characters.
\section{Conclusions and Future Work}
  Human beings react to each other's affective state and adjust their behaviors accordingly. In this paper we have proposed a method for generating facial behaviors in a dyadic interaction. Our model learns semantically meaningful facial behaviors such as head pose and movements of landmarks and generates appropriate and temporally smooth sequences of facial expressions. 
  
In our future work, we are interested in transferring the generated facial expressions to an avatar's face and generating sequences of expressive avatars reacting to the interviewee's affective state using the proposed approaches in this work. We then will run a user study using pairs of interviewee and generated avatar's videos to study the most appropriate responses for each video and will use the result in building our final models of emotionally intelligent agents. 

% \section{ACKNOWLEDGMENTS}

\bibliographystyle{abbrv}
\bibliography{egpaper_for_review.bib}
\end{document}